\documentclass[sigconf]{acmart}

\copyrightyear{2026}
\acmYear{2026}
\setcopyright{cc}
\setcctype{by}
\acmConference[WWW '26] {Proceedings of the ACM Web Conference 2026}{April 13--17, 2026}{Dubai, United Arab Emirates.}
\acmBooktitle{Proceedings of the ACM Web Conference 2026 (WWW '26), April 13--17, 2026, Dubai, United Arab Emirates}
\acmISBN{979-8-4007-2307-0/2026/04}
\acmDOI{10.1145/3774904.3792732}

\AtBeginDocument{%
  }

\usepackage{multirow}
\usepackage{subfigure}
\usepackage{booktabs}

\begin{document}

\title{MemWeaver: A Hierarchical Memory from  Textual Interactive Behaviors for Personalized Generation}

\author{Shuo Yu}
\orcid{0009-0006-1060-5451}
\affiliation{%
  \institution{State Key Laboratory of Cognitive Intelligence, University of Science and Technology of China}
  \city{Hefei}
  \state{}
  \country{China}
}
\email{yu12345@mail.ustc.edu.cn}

\author{Mingyue Cheng}
\authornote{Mingyue Cheng is the corresponding author.}
\orcid{0000-0001-9873-7681}
\affiliation{%
  \institution{State Key Laboratory of Cognitive Intelligence, University of Science and Technology of China}
  \city{Hefei}
  \state{}
  \country{China}
}
\email{mycheng@ustc.edu.cn}

\author{Daoyu Wang}
\orcid{0009-0002-0452-0516}
\affiliation{%
  \institution{State Key Laboratory of Cognitive Intelligence, University of Science and Technology of China}
  \city{Hefei}
  \state{}
  \country{China}
}
\email{wdy030428@mail.ustc.edu.cn}

\author{Qi Liu}

\orcid{0000-0001-6956-5550}
\affiliation{%
  \institution{State Key Laboratory of Cognitive Intelligence, University of Science and Technology of China}
  \city{Hefei}
  \state{}
  \country{China}
}
\email{qiliuql@ustc.edu.cn}

\author{Zirui Liu}
\orcid{0009-0002-7263-9607}
\affiliation{%
  \institution{State Key Laboratory of Cognitive Intelligence, University of Science and Technology of China}
  \city{Hefei}
  \state{}
  \country{China}
}
\email{liuzirui@mail.ustc.edu.cn}

\author{Ze Guo}
\orcid{0009-0006-9461-5007}
\affiliation{%
  \institution{State Key Laboratory of Cognitive Intelligence, University of Science and Technology of China}
  \city{Hefei}
  \state{}
  \country{China}
}
\email{gz1504921411@mail.ustc.edu.cn}

\author{Xiaoyu Tao}
\orcid{0009-0000-0634-6254}
\affiliation{%
  \institution{State Key Laboratory of Cognitive Intelligence, University of Science and Technology of China}
  \city{Hefei}
  \state{}
  \country{China}
}
\email{txytiny@mail.ustc.edu.cn}

\renewcommand{\shortauthors}{Shuo Yu et al.}

\begin{abstract}
The primary form of user-internet engagement is shifting from leveraging implicit feedback signals, such as browsing and clicks, to harnessing the rich explicit feedback provided by textual interactive behaviors. This shift unlocks a rich source of user textual history, presenting a profound opportunity for a deeper form of personalization. 
However, prevailing approaches offer only a shallow form of personalization, as they treat user history as a flat list of texts for retrieval and fail to model the rich temporal and semantic structures reflecting dynamic nature of user interests.
In this work, we propose \textbf{MemWeaver}, a framework that weaves the user's entire textual history into a hierarchical memory to power deeply personalized generation. The core innovation of our memory lies in its ability to capture both the temporal evolution of interests and the semantic relationships between different activities.  To achieve this, MemWeaver builds two complementary memory components that both integrate temporal and semantic information, but at different levels of abstraction: behavioral memory, which captures specific user actions, and cognitive memory, which represents long-term preferences. This dual-component memory serves as a comprehensive representation of the user, allowing large language models (LLMs) to reason over both concrete  behaviors and abstracted cognitive traits. This leads to content generation that is deeply aligned with their latent preferences.
Experiments on the six datasets of the Language Model Personalization (LaMP) benchmark validate the efficacy of MemWeaver.
Our code is available\footnote{\url{https://github.com/fishsure/MemWeaver}}.

\end{abstract}





\maketitle

\section{Introduction}

\begin{figure}
    \centering
    \includegraphics[width=1\linewidth]{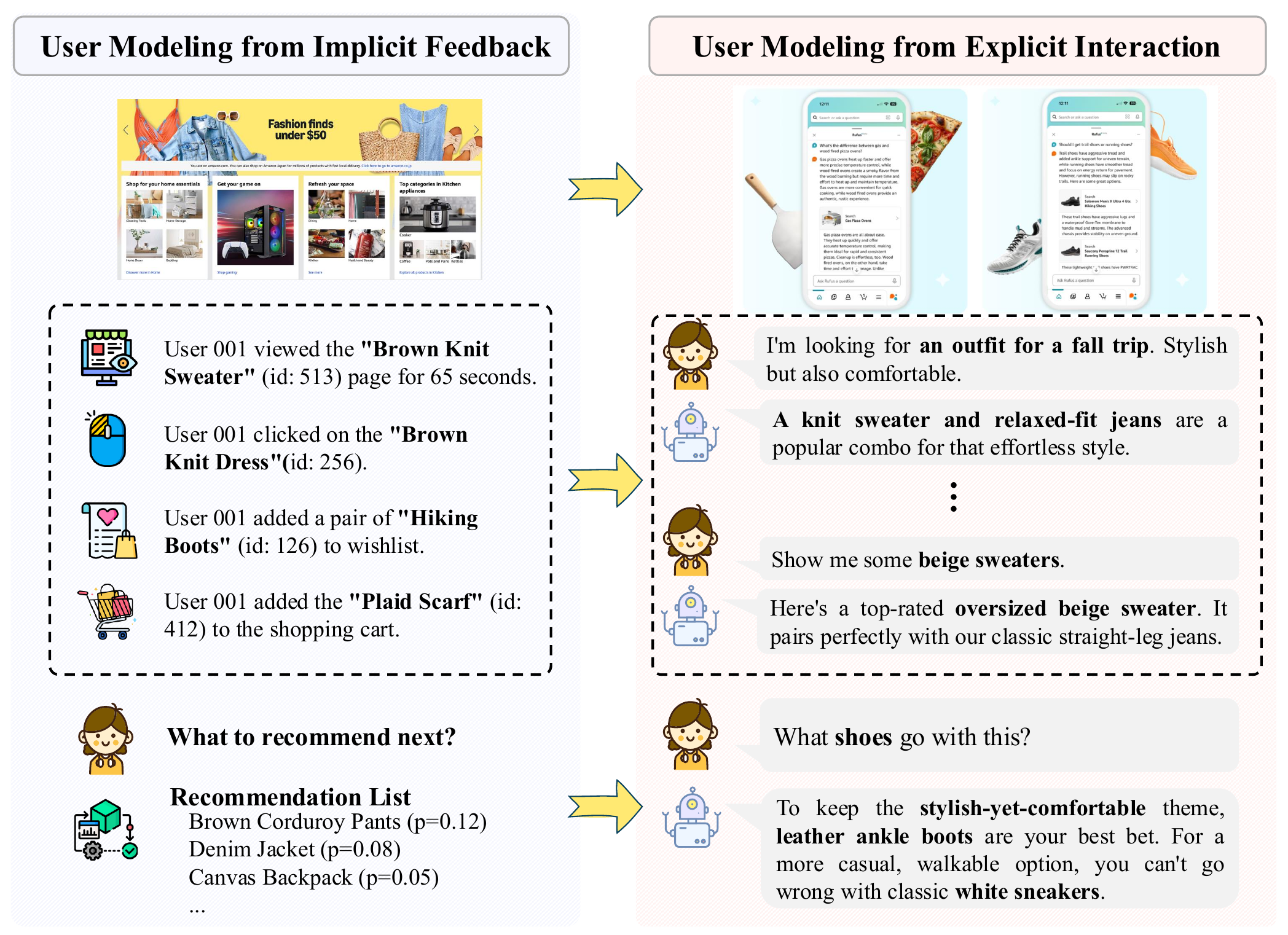}
\caption{
    \textbf{The diagram illustrates a key paradigm shift in user modeling, moving from leveraging implicit feedback signals like clicks to harnessing the rich, explicit feedback provided by user textual interactions.}
}
    \label{fig:enter-label}

\end{figure}

For decades, the dominant paradigm for personalization has been rooted in recommender systems that model user preferences from implicit feedback signals, such as clicks and purchases~\cite{hu2008collaborative,implicit,rec3,wang2025personalized,wang2025mitigating}. In this paradigm, users are often represented as vectors in a latent space, derived from their interactions with a catalog of item IDs\cite{www7_5,liu2022one}. However, a fundamental shift is underway, with the primary mode of user engagement evolving from passive content consumption guided by implicit behaviors to the active co-creation of information through explicit textual conversations~\cite{rec_survey,sun2019bert4rec,rec_llm}. As users increasingly engage in these direct interactive behaviors, the volume and depth of their textual history grow, creating a new foundation for deep user modeling and personalization~\cite{p5,www3_7}.

Despite this ongoing paradigm shift, much of the existing user modeling approach remains reliant on implicit signals~\cite{wang2019ngcf,he2020lightgcn}.  Classical approaches, such as collaborative filtering and matrix factorization, construct user profiles by learning embeddings from a history of interactions with atomic item identifiers (IDs)~\cite{kang2018sasrec}. While effective for capturing broad preference patterns from sparse data, these ID-based models have an inherent limitation: they are not designed to interpret the rich semantic content of language. The user's intent is learned indirectly through collaborative patterns, rather than being understood directly from their own words, leaving the nuanced details of their goals largely unexplored.

The emergence of this rich textual history creates a profound opportunity for a deeper form of personalization~\cite{pLLm_survey1,cheng2026mind2report}. This history offers a direct, high-fidelity narrative of a user's evolving interests, nuanced intentions, and specific preferences, far exceeding the expressive power of item IDs. However, harnessing this information is not trivial. The true value of this history lies not in its sheer volume, but in its intricate nature, embodying two critical dimensions: the temporal evolution of a user's interests and the latent semantic connections that link disparate topics~\cite{cfrag,ropg}. The central challenge, therefore, is to develop a framework capable of harnessing  both the temporal evolution and the semantic connections within this history to model the user.

To address this challenge, we introduce MemWeaver, a framework designed to weave a hierarchical memory from user interaction history. To capture both the temporal evolution of interests and the semantic relationships between user behaviors, MemWeaver constructs a dual-component memory comprising a behavioral memory that captures concrete actions and a cognitive memory that abstracts higher-level intent. To form the behavioral memory, we first represent the user's history as a network where each behavior is a node, interconnected by two types of edges: temporal edges link consecutive behaviors, while semantic edges connect those with high thematic relevance~\cite{graphrag,www2_22}.  The memory is then extracted from this network via a context-aware random walk~\cite{random_walk} that simulates the associative recall process of human memory, traversing the established edges. Concurrently, the cognitive memory is abstracted through a process that segments the history into distinct phases, summarizes the interests within each, and integrates these summaries into a global profile of user preferences.

Extensive experiments on the Language Model Personalization (LaMP) benchmark~\cite{lamp} validate our framework, demonstrating that our MemWeaver outperforms strong baselines. Our analysis confirms that a comprehensive user representation requires the joint modeling of temporal correlations to capture interest evolution and semantic relationships to link disparate concepts. Furthermore, we reveal a critical synergy between our dual memory components: the behavioral memory grounds the model in concrete actions, while the abstract cognitive memory provides the high-level guidance for coherent outputs, validating that this synergy between concrete context and abstract guidance is essential for achieving a deeper level of user alignment. 
Finally, our analysis of the incremental update mechanism confirms its efficiency and scalability. Experiments show that it achieves performance comparable to a full memory reconstruction at a fraction of the computational cost, making it suitable for continuous deployment.

In summary, the main contributions of this paper are as follows:
\begin{itemize}

\item We propose MemWeaver, a novel framework that weaves user history into a hierarchical memory, moving beyond the representational limitations of conventional flat logs.
\item We introduce a methodology to construct this memory, featuring an extracted behavioral memory to capture concrete context and an  abstracted cognitive memory to represent a user's evolving preferences.
\item Extensive experiments on the LaMP benchmark validate our framework, establishing state-of-the-art performance and confirming the synergistic effect of the dual-memory components through in-depth analysis.
\end{itemize}

\begin{figure*}
    \centering
    \includegraphics[width=0.98\linewidth]{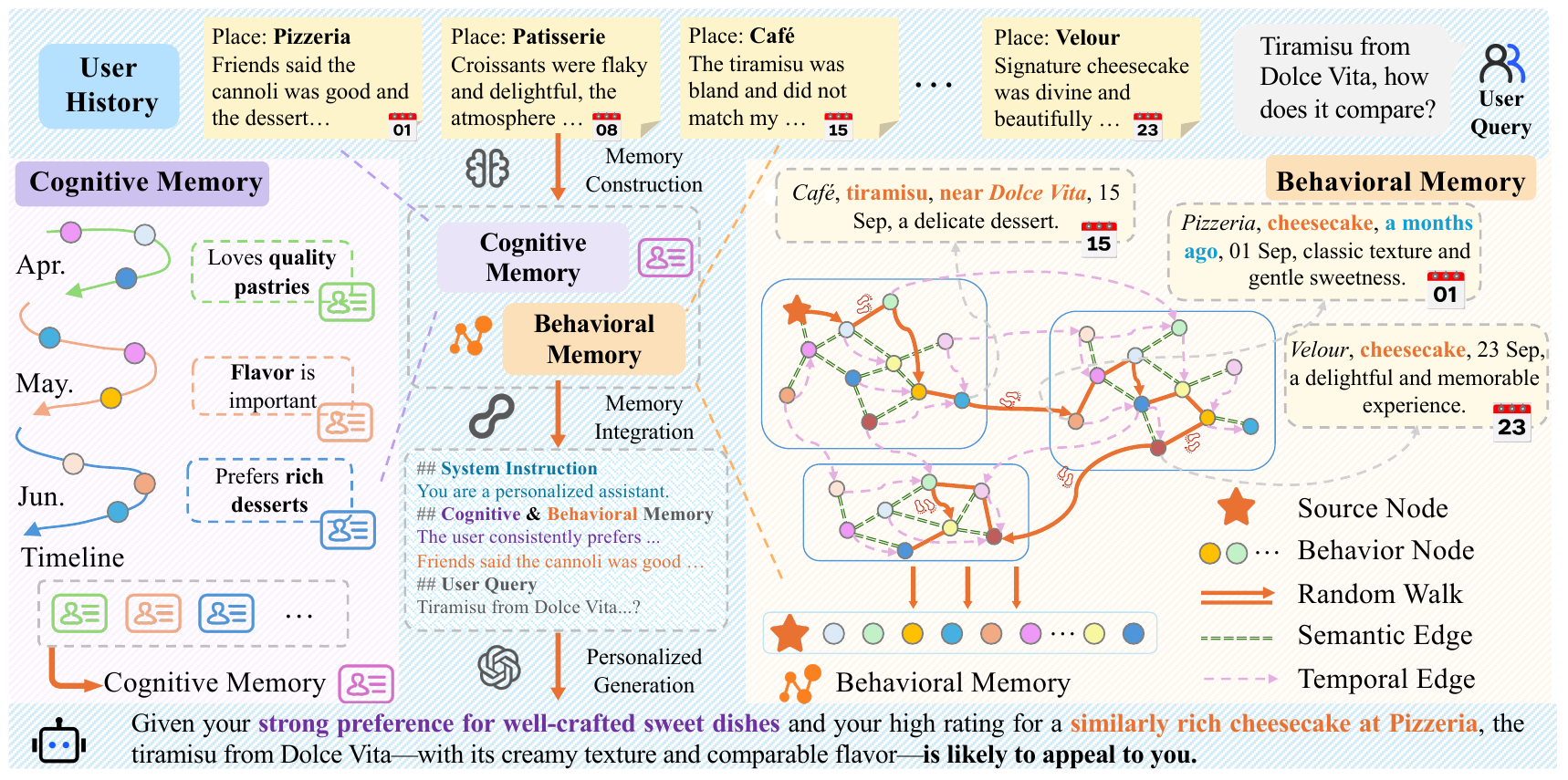}
\caption{An overview of the MemWeaver framework. MemWeaver builds a dual memory from user history logs: a \textbf{cognitive memory} for evolving preferences and a \textbf{behavioral memory} for capturing specific user behavior. Both memories are then structured into a comprehensive prompt to guide an LLM for personalized generation.}
    \label{fig:framework}
\end{figure*}
\section{Related Work}

\subsection{User Modeling for Personalization}

\label{sec:related}
Traditional personalization methods evolved from ID-based collaborative filtering and matrix factorization~\cite{koren2009matrix, rendle2010bpr,liuelo,liu2024computerized} to modeling with RNNs and Transformers~\cite{kang2018sasrec, sun2019bert4rec} and graph-based models~\cite{wang2019ngcf,he2020lightgcn}. 
 Despite increasing improvement in modeling temporal patterns and relational structures, these methods share a fundamental limitation: they are tailored for sparse, ID-level signals and are ill-equipped to interpret the rich semantics of textual interactions~\cite{rec5,zhang2023towards}. The advent of LLMs introduces a new frontier, shifting the focus to directly understanding user narratives from textual history via retrieval-augmented generation (RAG)~\cite{lewis2020rag,www6_6,yu2025multi}. However, the efficacy of current RAG-based personalization is hampered by simplistic, single-dimensional retrieval heuristics, such as recency-based or similarity-based approaches~\cite{llm2,karpukhin2020dpr, reimers2019sentencebert}, and their treatment of retrieved context as a flat, unstructured list of text chunks. 
 % This approach fails to fully exploit the LLM's reasoning capabilities, as it strips away the intricate relationships within a user's history. A critical gap therefore lies in the absence of a framework capable of constructing and leveraging a hierarchical memory from textual interaction histories.
Discarding relational context fails to fully exploit LLM reasoning, revealing a critical gap: the lack of a framework to derive hierarchical memory from textual interaction histories.

\subsection{Memory-Augmented LLMs} \label{sec:memory} Recent studies have investigated memory mechanisms to boost LLMs in long-context understanding, multi-turn reasoning, and personalization~\cite{memory_survey, mem0, memorybank}. To facilitate long-context modeling, prior works either extend context windows or compress information: MemoRAG~\cite{memorag} and Memory$^3$~\cite{memory3} apply KV-compression, while Mem-Agent~\cite{memagent} and Memory-R1~\cite{memoryr1} utilize reinforcement learning to refine summarization policies that preserve salient information. For complex reasoning, memory modules support multi-hop and agentic decision-making; for instance, HippoRAG~\cite{hipporag} emulates associative reasoning via knowledge graphs, while Memento and Agent-KB~\cite{agentkb,wang2025paperarena} store past experiences for in-context reuse. Further, Mirix~\cite{mirix} and Mem-OS~\cite{memos} integrate heterogeneous memory types to support general-purpose agent behaviors. For personalization, Prime~\cite{prime} employs a dual-memory structure with chain-of-thought, whereas MiLP~\cite{milp} uses Bayesian-optimized LoRA modules as parametric memory to capture user-specific signals. Despite these advances, existing methods often fail to jointly model the intertwined semantic and temporal structures in user histories. To address this limitation, we propose MemWeaver, a framework that constructs a hierarchical memory to capture both the semantic and temporal information within user history.
\section{Methodology}
In this section, we present our MemWeaver framework. We first formulate the problem of personalized text generation, then detail the construction of our dual-memory system, and finally explain their integration to guide the personalized generation process.

\subsection{Problem Formulation}

Let $\mathcal{U} = \{u_1, u_2, \ldots, u_M\}$ denote the set of users, where $M$ is the total number of users. 
Each user $u \in \mathcal{U}$ is associated with a chronologically ordered history $\mathcal{H}_u = [d_1, d_2, \ldots, d_{N_u}]$, consisting of $N_u$ documents that the user has previously authored or interacted with. 
Formally, the personalized text generation dataset is defined as $\mathcal{D} = \{(u, \mathcal{H}_u, q, y)_i\}_{i=1}^{|\mathcal{D}|}$, where $q$ is a user-issued query and $y$ is the corresponding ground-truth response.
The goal of personalized generation is to model the probability of a response $y'$ conditioned on the user's history $\mathcal{H}_u$ and the current query $q$. Formally, the generation process can be defined as:
\begin{equation}
\hat{y} = \arg\max_{y'} p(y' \mid q, \mathcal{H}_u; \theta),
\label{eq:personalized_generation}
\end{equation}
where $\theta$ denotes the parameters of the LLM.

\subsection{Principles of Memory Design}

To address the challenges of leveraging long and unstructured user histories as defined in Equation~\ref{eq:personalized_generation}, we propose \textbf{MemWeaver}, a cognition-inspired framework designed to organize and interpret user history. As detailed in Figure \ref{fig:framework} , this framework converts the raw history $\mathcal{H}_u$ into a structured, dual-component memory system designed to provide complementary signals for personalization. 
Inspired by human cognition, MemWeaver constructs two distinct memory representations: the behavioral memory, $\mathcal{M}^{\mathrm{behavior}}_u \subseteq \mathcal{H}_u$, a dynamically extracted subset of the user's history that provides fine-grained, query-specific context; and the cognitive memory, $\mathcal{M}^{\mathrm{cog}}_u$, a natural language summary that abstracts the user's long-term, evolving preferences.
With these components, our framework reframes the original generation task. Instead of conditioning on the entire, unwieldy history $\mathcal{H}_u$, the framework is conditioned on our structured memory representations. Specifically, the final response is generated by jointly conditioning the LLM on the query, the behavioral memory, and the cognitive memory constructed from user behaviors to achieve superior personalization.

\subsection{Memory Construction and Updating}

\subsubsection{Behavioral Memory Extraction}

 To provide a compact and contextually relevant representation for a given query, our behavioral memory, $\mathcal{M}^{\mathrm{behavior}}_u$, is an extracted subset of a user's history. The construction process involves first building a structured representation of the user's history and then performing a guided search to extract the memory.
To begin, we create the memory graph that serves as the foundational structure. Given a user $u$ and their behavioral history $\mathcal{H}_u = [d_1, d_2, \ldots, d_{N_u}]$ ordered by time, we first encode each instance $d_i$ into a dense vector representation $e_i \in \mathbb{R}^d$ using a pre-trained sentence-level transformer (e.g., BGE-M3~\cite{bge_embedding}). To distill latent thematic structures from this history, we then employ the $K$-means clustering algorithm~\cite{kmeans} on the behavior embeddings. This process partitions the user's behaviors into a set of $K$ distinct semantic clusters, $\{\mathcal{C}_1, \ldots, \mathcal{C}_K\}$, where each cluster represents a coherent thematic interest. Based on this, we construct a personalized memory graph $\mathcal{G}_u = (\mathcal{V}, \mathcal{E})$, where each node $v_i \in \mathcal{V}$ corresponds to a behavior $d_i$. Two types of undirected edges are defined: temporal edges $(v_i, v_{i+1})$ link consecutive behaviors, preserving sequential integrity, while semantic edges $(v_i, v_j)$ connect any two nodes within the same semantic cluster $\mathcal{C}_k$, capturing thematic relations.

With this structured memory graph in place, the objective is to extract a relevant subset of nodes conditioned on a given query. A naive retrieval of all nodes is infeasible for long histories, while simple heuristics may fail to leverage the rich graph structure. Therefore, we employ a context-aware random walk, a technique conceptually inspired by the associative nature of human memory recall, where one thought triggers a chain of related recollections. This probabilistic approach is not only powerful for exploring complex graph structures but also allows us to efficiently sample a coherent and contextually relevant path of behaviors, which forms the behavioral memory,  guiding the final response generation.

At each step, the walk transitions from a node $u$ to a neighbor $v$ based on a score $S(u \rightarrow v)$. This score combines semantic relevance with two temporal biases: a recency score $R(v) = \exp(-\lambda_1 \Delta t_v)$ and a continuity score $C(u, v) = \exp(-\lambda_2 \Delta t_{uv})$. Therefore, the final transition score is calculated as:
\begin{equation}
S(u \rightarrow v) = \left( \frac{e_q \cdot e_v}{\|e_q\| \|e_v\|} \right)^{\alpha} \cdot R(v) \cdot C(u, v),
\label{eq:score}
\end{equation}
where the hyperparameters $\alpha, \lambda_1,$ and $\lambda_2$ control the influence of semantic relevance, the penalty for the time gap from the query ($\Delta t_v$), and the penalty for sequential distance ($\Delta t_{uv}$), respectively. 
The likelihood of transitioning to a neighboring node is directly proportional to this score. Specifically, the transition probabilities are calculated by normalizing these scores over all potential neighbors. The walk proceeds according to this probability distribution until a predefined step limit is reached, forming the behavioral memory, $\mathcal{M}^{\mathrm{behavior}}_u$, from the sequence of unique nodes visited.

\subsubsection{Cognitive Memory Abstraction}

While the behavioral memory captures transient, query-specific signals, the hierarchical cognitive memory, $\mathcal{M}^{\mathrm{cog}}_u$, is designed to distill a longitudinal and abstract representation of the user's stable preferences. This is achieved through a hierarchical abstraction process that first partitions the user's history into meaningful temporal phases and then synthesizes these phases into a cohesive, narrative-style summary.
Instead of treating the user's history as a monolithic sequence, we first partition their chronologically ordered behaviors, $\mathcal{H}_u$, into a series of temporally coherent segments, $\{\mathcal{H}_1, \mathcal{H}_2, \ldots, \mathcal{H}_T\}$. Each segment represents a distinct phase of the user's interests. The boundaries of these segments are determined by identifying semantic breakpoints in the user's history—points where the semantic similarity between consecutive behaviors drops significantly, indicating a potential shift in the user's focus. This segmentation is further governed by rule-based constraints to ensure that each resulting segment, $\mathcal{H}_t$, is meaningfully sized, preventing them from being too granular to lose context or too broad to conflate distinct topics.

With the user history partitioned into meaningful phases, we employ a two-stage summarization process using an LLM to construct the cognitive memory.
The process begins at a granular level, where the LLM processes each temporal segment to generate a local summary, $s_t$, that abstracts the user's interests during that specific time period. This is formulated for all segments $t=1, \ldots, T$ as:
\begin{equation}
s_t = \text{LLM}(\mathcal{H}_t),
\end{equation}
in which $\mathcal{H}_t$ represents the corresponding segment of the user's historical behaviors. These local summaries are then collected into a set, $S=\{s_1, s_2, \ldots, s_T\}$, which serves as the input for a higher-level synthesis. In this synthesis step, the LLM functions as a preference integrator, tasked with weaving together these distinct thematic nuclei. It identifies overarching themes, establishes connections between different interest phases, and articulates the evolution or migration of the user's preferences over time. This process results in a single, cohesive global summary that constitutes the user's cognitive memory, which is generated as $\mathcal{M}^{\mathrm{cog}}_u = \text{LLM}(S)$, providing a narrative-style representation of the user's long-term preferences.

\begin{table*}[t]
\centering

\caption{
Comparison of the performance of MemWeaver with other approaches on the LaMP benchmark. 
$\uparrow$ indicates that a higher value is better, while $\downarrow$ indicates that a lower value is better. 
The best results are in bold and the second best are underlined.
All reported improvements of MemWeaver over baselines are statistically significant (t-tests, $p$-value $< 0.05$).
}
\label{tab:main_result}
\resizebox{.98\textwidth}{!}{
% The | characters here create the vertical lines after the first two columns.
\begin{tabular}{l|l|cccccccccccc} 
\toprule
\multicolumn{1}{l}{\multirow{2}{*}{LLMs}} & 
\multicolumn{1}{l}{\multirow{2}{*}{Methods}} & 
\multicolumn{2}{c}{LaMP-1} & 
\multicolumn{2}{c}{LaMP-2} & 
\multicolumn{2}{c}{LaMP-3} & 
\multicolumn{2}{c}{LaMP-4} & 
\multicolumn{2}{c}{LaMP-5} & 
\multicolumn{2}{c}{LaMP-7} \\
\cmidrule(l){3-4} \cmidrule(l){5-6} \cmidrule(l){7-8} 
\cmidrule(l){9-10} \cmidrule(l){11-12} \cmidrule(l){13-14}
\multicolumn{1}{c}{}  & \multicolumn{1}{c}{} 
&Acc.~$\uparrow$ &F1~$\uparrow$ &Acc.~$\uparrow$ &F1~$\uparrow$ &MAE~$\downarrow$ &RMSE~$\downarrow$ &R-1~$\uparrow$ &R-L~$\uparrow$ &R-1~$\uparrow$ &R-L~$\uparrow$ &R-1~$\uparrow$ &R-L~$\uparrow$ \\
\midrule
\multirow{8}*{Qwen3}
&Vanilla &0.5200 &0.2600 &0.3766 &0.0251 &0.5200 &0.8067 &0.1332 &0.1154 &0.4443 &0.3801 &0.4281 &0.3767 \\
&Random &0.5800 &0.2900 &0.3845 &0.0256 &0.3968 &0.5433 &0.1580 &0.1269 &0.4537 &0.3984 &0.4395 &0.3738 \\
&Recency &0.6400 &0.3200 &0.3933 &0.0262 &0.3665 &0.6064 &0.1649 &0.1424 &0.4559 &0.3959 &0.4468 &0.3794 \\
&BM25 &0.6476 &0.3238 &0.3879 &0.0257 &0.3515 &0.5201 &0.1645 &0.1477 &0.4584 &0.3978 &0.4555 &0.3842 \\
&BGE &0.6498 &0.3249 &0.3933 &0.0262 &0.3433 &0.4900 &0.1637 &0.1466 &0.4696 &0.4059 &0.4618 &0.3902 \\
&ROPG &0.6516 &0.3258 &0.4254 &0.0278 &0.3315 &0.4734 &0.1634 &0.1457 &0.4648 &0.3998 &0.4622 &0.3929 \\
&CFRAG &\underline{0.6566} &\underline{0.3283} &\underline{0.4400} &\underline{0.0288} &\underline{0.3266} &\underline{0.4600} &\underline{0.1678} &\underline{0.1492} &\underline{0.4698} &\underline{0.4061} &\underline{0.4646} &\underline{0.3955} \\
&\textbf{MemWeaver} &\textbf{0.6733} &\textbf{0.3367} &\textbf{0.4633} &\textbf{0.0311} &\textbf{0.2800} &\textbf{0.3733} &\textbf{0.1724} &\textbf{0.1544} &\textbf{0.4757} &\textbf{0.4138} &\textbf{0.4792} &\textbf{0.4079} \\
\midrule % <-- This was \hline, now changed to \midrule to create the broken line effect.

\multirow{8}*{Llama3.1}
&Vanilla  &0.4933 &0.2467 &0.3567 &0.0251 &0.4501 &0.6633 &0.1313 &0.1167 &0.4407 &0.3504 &0.3341 &0.2564 \\
&Random &0.5233 &0.2617 &0.4300 &0.0275 &0.3300 &0.4567 &0.1451 &0.1295 &0.4451 &0.3619 &0.3204 &0.2625 \\
&Recency &0.5500 &0.2750 &0.4431 &0.0295 &0.2914 &0.3662 &0.1658 &0.1517 &0.4513 &\underline{0.4013} &0.3294 &0.2660 \\
&BM25 &0.5533 &0.2767 &0.4500 &0.0301 &0.2700 &\underline{0.3366} &0.1743 &0.1604 &0.4512 &0.3882 &\underline{0.3396} &0.2729 \\
&BGE 
&0.5602 &0.2801 &0.4583 &0.0304 &0.2733 &0.3600 &0.1713 &0.1570 &0.4560 &0.3919 &0.3236 &0.2643 \\
&ROPG
&0.5894 &0.2947 &0.4662 &0.0311 &0.2694 &0.3512 &0.1798 &0.1684 &0.4578 &0.3955 &0.3304 &0.2671 \\
&CFRAG &\underline{0.6145} &\underline{0.3073} &\underline{0.4681} &\underline{0.0312} &\underline{0.2630} &0.3451 &\underline{0.1821} &\underline{0.1702} &\underline{0.4614} &\underline{0.4013} &0.3389 &\underline{0.2781} \\
&\textbf{MemWeaver} &\textbf{0.6533} &\textbf{0.3267} &\textbf{0.4733} &\textbf{0.0315} &\textbf{0.2533} &\textbf{0.3300} &\textbf{0.1899} &\textbf{0.1794} &\textbf{0.4718} &\textbf{0.4138} &\textbf{0.3581} &\textbf{0.2916} \\
\bottomrule
\end{tabular} 
}
\end{table*}

\subsubsection{Incremental Memory Updating}
To ensure our framework can efficiently handle continuously growing user histories without costly full reconstructions, we employ lightweight incremental update mechanisms for both memory components. For the behavioral memory, when a new batch of behaviors $\mathcal{H}_{\text{new}}$ arrives, it is encoded and integrated without reprocessing the existing graph. Temporal continuity is maintained by linking the last existing node to the first new one, and new semantic edges are established by clustering the new behaviors independently. Similarly, for the cognitive memory, the new batch is processed independently by partitioning it into coherent segments and generating new local summaries. The global cognitive memory is then re-synthesized by prompting the
LLM with the full collection of local summaries, combining the previously generated ones 
with the new ones. This approach requires only a single, high-level integration step to update the user's overall narrative, making the process computationally tractable.
Operating on condensed summaries bounds token costs as history grows, enabling real-time adaptation to evolving interests without the latency of full-context re-computation.

\subsection{Memory-Augmented  Generation}

Based on the behavioral and cognitive memory, our framework then guides the LLM's generation process by integrating both components.  The overall objective is to generate a response $\hat{y}$ that maximizes the conditional probability given the query, the behavioral memory and cognitive memory:
\begin{equation}
\hat{y} = \arg\max_{y'} p\big(y' \mid q, \mathcal{M}^{\mathrm{behavior}}_u, \mathcal{M}^{\mathrm{cog}}_u; \theta\big),
\label{eq:final_gen}
\end{equation}
where $\theta$ represents the parameters of the LLM.
To achieve this, the two memory components provide complementary signals. The behavioral memory, $\mathcal{M}^{\mathrm{behavior}}_u$, offers concrete examples, grounding the generation in specific user actions and short-term context. Concurrently, the cognitive memory, $\mathcal{M}^{\mathrm{cog}}_u$, serves as a high-level instruction defining the user's long-term preferences to ensure global consistency. This joint conditioning on both concrete examples and abstract guidance allows our framework to balance short-term relevance with long-term user consistency, producing deeply personalized responses.
\section{Experiments}
To validate the overall effectiveness of MemWeaver and the contributions of its core components, we conduct comprehensive experiments, including main comparisons against strong baselines and detailed ablation studies.
\subsection{Experimental Setup}
% Our experimental setup is as follows: we first introduce the LaMP benchmark dataset, then specify the evaluation metrics for each task, and finally describe the baseline methods used for comparison.

% Our experimental setup is as follows: we introduce the LaMP benchmark and the evaluation metrics, and finally describe the baseline methods used for comparison.
\subsubsection{Dataset} 

We adopt the LaMP benchmark~\cite{lamp}, which covers seven personalized tasks, including three classification tasks—\textbf{LaMP-1} (Citation Identification, binary), \textbf{LaMP-2} (Movie Tagging, multi-class with 15 categories), and \textbf{LaMP-3} (Product Rating, ordinal from 1–5 stars)—and four text generation tasks: \textbf{LaMP-4} (News Headline Generation), \textbf{LaMP-5} (Scholarly Title Generation), \textbf{LaMP-6} (Email Subject Generation), and \textbf{LaMP-7} (Tweet Paraphrasing). Since LaMP-6 is not publicly available, we exclude it from our experiments. Following the official protocol, we adopt a time-based split, where each user’s interactions are partitioned into training, validation, and test sets by timestamp. This setting simulates a realistic  scenario and allows us to study the effect of recency. To demonstrate the generalizability of our approach, we conduct additional experiments on other datasets which are detailed in Appendix \ref{sec:appendix_exp}.

\begin{table*}[t]
\centering
\caption{Comprehensive ablation study of MemWeaver at both the main module and fine-grained component levels.}
\label{tab:comprehensive_ablation}
\resizebox{0.98\textwidth}{!}{
% The preamble has been updated to add vertical lines for clarity
\begin{tabular}{l|l|cccccccccccc}
\toprule
% The header "Method" is changed to "Configuration"
\multicolumn{1}{c|}{\multirow{2}{*}{Component}} & \multicolumn{1}{l|}{\multirow{2}{*}{Configuration}} & \multicolumn{2}{c}{LaMP-1} & \multicolumn{2}{c}{LaMP-2} & \multicolumn{2}{c}{LaMP-3} & \multicolumn{2}{c}{LaMP-4} & \multicolumn{2}{c}{LaMP-5} & \multicolumn{2}{c}{LaMP-7} \\
\cmidrule(lr){3-4} \cmidrule(lr){5-6} \cmidrule(lr){7-8} \cmidrule(lr){9-10} \cmidrule(lr){11-12} \cmidrule(lr){13-14}
\multicolumn{1}{c|}{} & \multicolumn{1}{c|}{} & {Acc.~$\uparrow$} & {F1~$\uparrow$} & {Acc.~$\uparrow$} & {F1~$\uparrow$} & {MAE~$\downarrow$} & {RMSE~$\downarrow$} & {R-1~$\uparrow$} & {R-L~$\uparrow$} & {R-1~$\uparrow$} & {R-L~$\uparrow$} & {R-1~$\uparrow$} & {R-L~$\uparrow$} \\
\midrule
% The "MemWeaver" row is now split into two columns for consistency
\textbf{Full Model} & \textbf{MemWeaver} &\textbf{0.6733} &\textbf{0.3367} &\textbf{0.4633} &\textbf{0.0311} &\textbf{0.2800} &\textbf{0.3733} &\textbf{0.1724} &\textbf{0.1544} &\textbf{0.4757} &\textbf{0.4138} &\textbf{0.4792} &\textbf{0.4079} \\
\midrule
\multirow{2}{*}{\begin{tabular}{@{}c@{}}Main\\Component\end{tabular}} & w/o Cognitive Memory & 0.6417 & 0.3209 & 0.4525 & 0.0301 & 0.2988 & 0.3871 & 0.1651 & 0.1477 & 0.4630 & 0.4076 & 0.4543 & 0.3817 \\
& w/o Behavioral Memory & 0.6402 & 0.3401 & 0.4566 & 0.0286 & 0.6400 & 1.2500 & 0.1069 & 0.0964 & 0.4209 & 0.3499 & 0.4279 & 0.3556 \\
\midrule
\multirow{3}{*}{\begin{tabular}{@{}c@{}}w/ Behavioral\\Memory\end{tabular}} & w/o Temporal Edges & 0.6233 & 0.3117 & 0.4417 & 0.0291 & 0.3016 & 0.4084 & 0.1636 & 0.1461 & 0.4621 & 0.4032 & 0.4488 & 0.3771 \\
& w/o Semantic Edges & 0.6324 & 0.3162 & 0.4483 & 0.0297 & 0.2933 & 0.3900 & 0.1612 & 0.1448 & 0.4560 & 0.3919 & 0.4436 & 0.3743 \\
& w/o Edge Weighting & 0.5708 & 0.2854 & 0.4533 & 0.0301 & 0.3300 & 0.4733 & 0.1614 & 0.1440 & 0.4620 & 0.3995 & 0.4431 & 0.3701 \\
\midrule
\multirow{2}{*}{\begin{tabular}{@{}c@{}}w/ Cognitive\\Memory\end{tabular}} & w/o Clustering & 0.6431 & 0.3216 & 0.4546 & 0.0302 & 0.2892 & 0.3710 & 0.1665 & 0.1487 & 0.4625 & 0.4052 & 0.4620 & 0.3893 \\
& w/o Global Summary & 0.6548 & 0.3274 & 0.4582 & 0.0304 & 0.2944 & 0.3725 & 0.1653 & 0.1480 & 0.4676 & 0.4089 & 0.4519 & 0.3805 \\
\bottomrule
\end{tabular}}
\end{table*}

\subsubsection{Evaluation Metrics}
As in prior work~\cite{lamp,ropg,cfrag}, we report accuracy(Acc.) and F1 score for LaMP-1 and LaMP-2, mean absolute error (MAE) and root mean squared error (RMSE) for LaMP-3, and for the generation tasks, ROUGE-1(R-1) and ROUGE-L(R-L)~\cite{rouge} for LaMP-4, LaMP-5 and LaMP-7.

\subsubsection{Baselines}
To assess the effectiveness of MemWeaver, we compare it against a range of baselines. These include \textit{Vanilla}, a non-personalized approach that feeds the query directly into the LLM without any user history. For personalized settings, we consider heuristic methods such as \textit{Random} (selecting $k$ random documents) and \textit{Recency} (selecting the $k$ most recent documents), as well as standard retrievers like BM25~\cite{bm25} and BGE~\cite{bge_embedding}. We further include advanced methods: \textit{ROPG}~\cite{ropg}, which trains the retriever via contrastive learning, using feedback from LLM generations, and \textit{CFRAG}~\cite{cfrag}, which improves retrieval by incorporating documents that were useful to users with similar preferences.

\subsubsection{Implementation Details}
For all experiments, we use Qwen3-8B~\cite{qwen3} and Llama-3.1-8B-Instruct~\cite{llama3} as the backbone LLMs to evaluate our framework and all compared baselines. We utilize BGE-M3~\cite{bge_embedding} as the retrieval model. Unless otherwise specified, results are reported using Qwen3-8B. Further implementation details are provided in Appendix \ref{sec:implementation_details}.

% \subsubsection{Implementation Details}

\subsection{Experimental Results}

As presented in Table \ref{tab:main_result}, our proposed framework, MemWeaver, demonstrates unwavering superiority by securing the top performance on all twelve evaluation metrics across all six LaMP benchmark datasets. These statistically significant improvements are consistent across all task types, including classification and text generation. This consistent outperformance can be attributed to MemWeaver's unique dual-memory architecture. While strong baselines effectively leverage semantic relevance, they primarily operate on a flat set of historical texts and struggle to capture the evolution of user preferences. MemWeaver, by contrast, explicitly models both dimensions: the cognitive memory synthesizes a long-term understanding of the user's core tastes, while the dynamically constructed behavioral memory captures the immediate context relevant to the current query. Furthermore, the consistent outperformance of advanced dense retrieval methods over simpler heuristics like BM25 and Recency underscores the critical role of deep semantic understanding in this domain.

Meanwhile, the consistent advantage of Recency over random sampling validates that temporal relevance remains a vital component. This finding, combined with the previously established importance of deep semantic understanding, suggests that state-of-the-art personalization requires the effective integration of both semantic and temporal signals. Therefore, MemWeaver is designed to construct its memory by explicitly drawing from both the semantic and temporal dimensions. Instead of simply retrieving a set of semantically similar or recent memories, it constructs a holistic user representation that is simultaneously grounded in stable, long-term interests and responsive to immediate, evolving context, leading to its superior performance.

\subsection{Ablation Study}

Table~\ref{tab:comprehensive_ablation} presents a comprehensive ablation study that systematically analyzes the contributions of MemWeaver's components, confirming their individual importance and collaborative interplay.

\subsubsection{Behavioral versus Cognitive Components}
Our primary analysis investigates the roles of MemWeaver's two core modules, the behavioral memory and the cognitive memory. The results confirm that both are indispensable, as removing either significantly degrades performance. More importantly, the study reveals a clear synergistic and hierarchical relationship. We identify behavioral memory as the foundational layer, as it provides the model with concrete, in-context examples from the user's actual history. Its absence causes a catastrophic performance collapse on context-heavy generative tasks such as LaMP-4. In contrast, the cognitive memory functions as a refinement layer, building upon this foundation to distill long-term, global preferences that refine the output. The comparatively smaller performance drop upon its removal supports this hierarchical view. To summarize, behavioral memory provides the essential context, while the cognitive memory enriches it with a global perspective for optimization.
\begin{figure}
    \centering
    \includegraphics[width=0.98\linewidth]{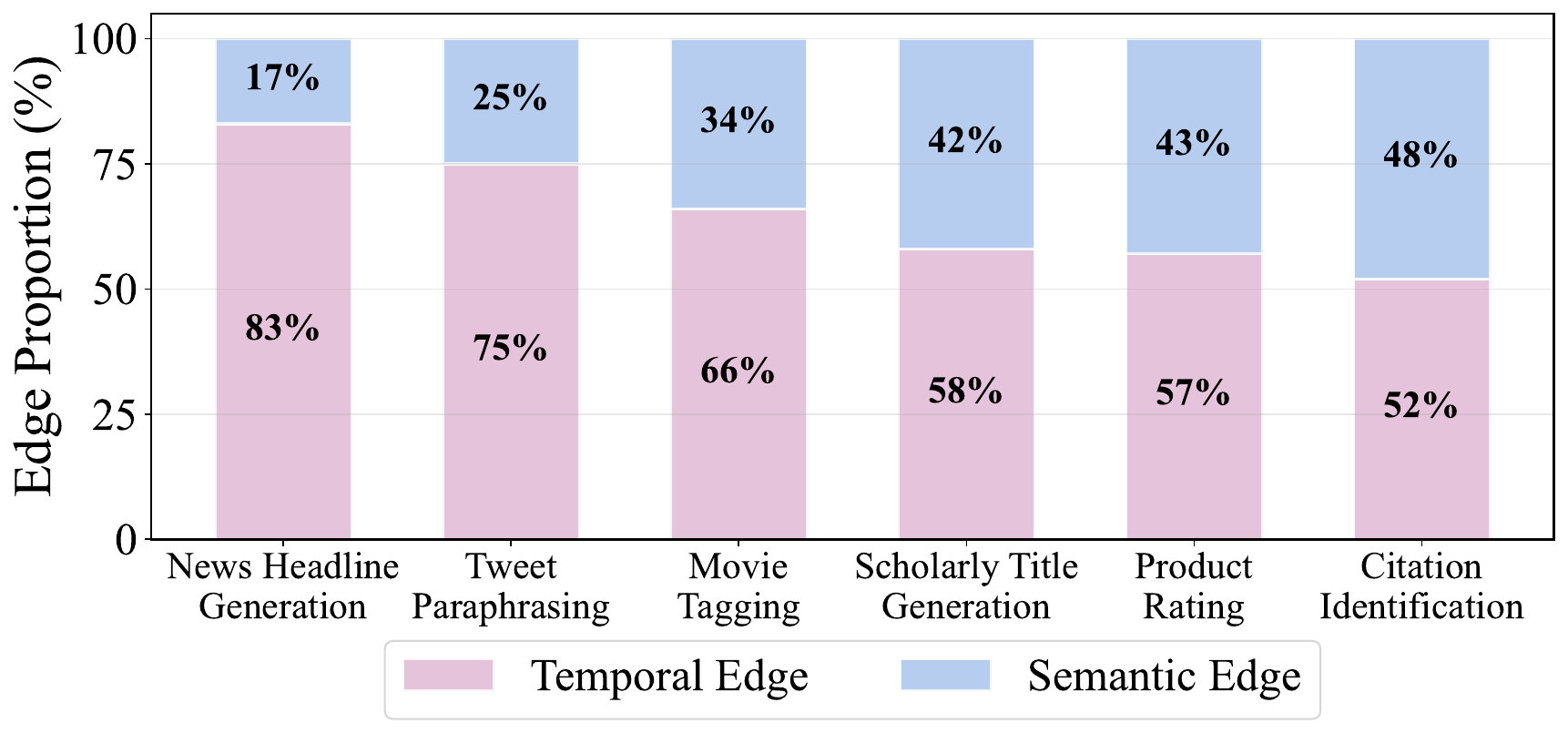}
\caption{Proportion of semantic and temporal edges in the behavioral memory across different LaMP datasets.}
\label{fig:edge_proportion}

\end{figure}

\begin{figure*}
    \centering
    \includegraphics[width=0.98\linewidth]{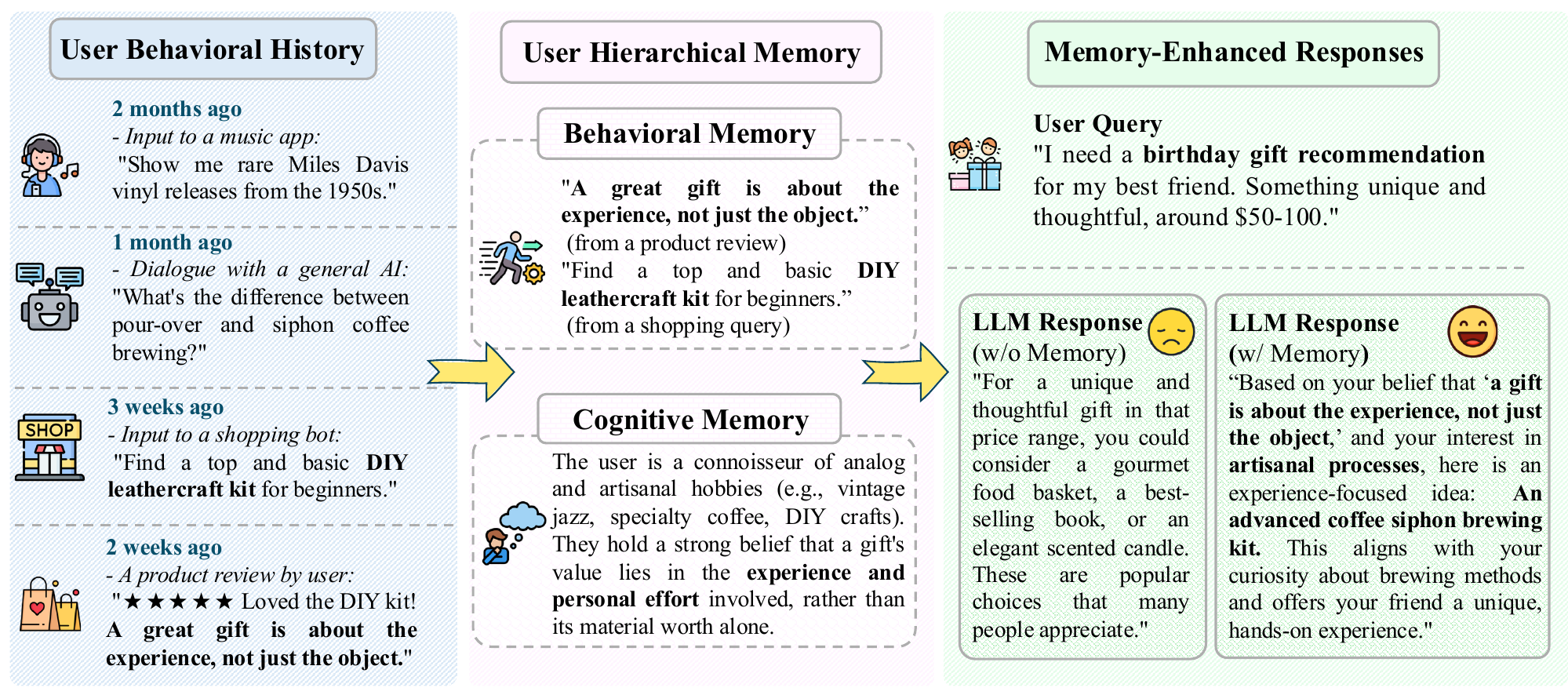}
\caption{An illustrative case study of MemWeaver. The framework transforms user history into a dual behavioral and cognitive memory, enabling an LLM to produce a personalized response that is far superior to the generic, memory-less baseline.}
    
    \label{fig:case_study}
\end{figure*}

\subsubsection{Analysis of Behavioral Memory Components}
We performed a fine-grained ablation to dissect the contributions of the components within the behavioral memory. The complete model consistently outperforms all its ablated variants, confirming that temporal edges, semantic edges, and the edge weighting mechanism are all integral to its performance. When comparing the two types of connections, we find that the removal of semantic edges causes a more significant performance drop than the removal of temporal edges. This suggests that the semantic content and relationships between user actions carry more weight in constructing an effective user profile than their purely sequential ordering. Finally, the study reveals that the edge weighting mechanism is the most critical component. The variant without edge weighting suffers the most severe performance degradation, demonstrating that simply constructing a graph of behaviors is insufficient. Instead, the ability to  quantify these relationships through fine-grained weighting is  important.

\subsubsection{Analysis of Cognitive Memory Strategies}
To investigate the contributions of the components within the cognitive memory, we performed a fine-grained ablation study. Findings underscore the importance of our hierarchical methodology. Removing the initial segmentation and clustering component leads to a significant performance degradation. This underscores the benefits of our approach: by partitioning the user's history into temporally coherent phases and summarizing them individually, our model can capture the distinct characteristics of the user's interests at different time periods. This staged process allows the final cognitive memory not only to achieve high-level semantic abstraction but also to effectively model the evolution of the user's interests over time. Furthermore, it avoids the pitfalls of summarizing the entire user history at once, which could lead to an excessively long context that degrades the quality of the resulting summary.  Results also confirm the efficacy of the final aggregation step, as the model variant without a global summary shows a consistent drop in performance, demonstrating that effectively distilling these temporal summaries into a single, high-level profile is critical for capturing the user's core traits.

\begin{figure}[t]
    \centering
    \includegraphics[width=0.98\linewidth]{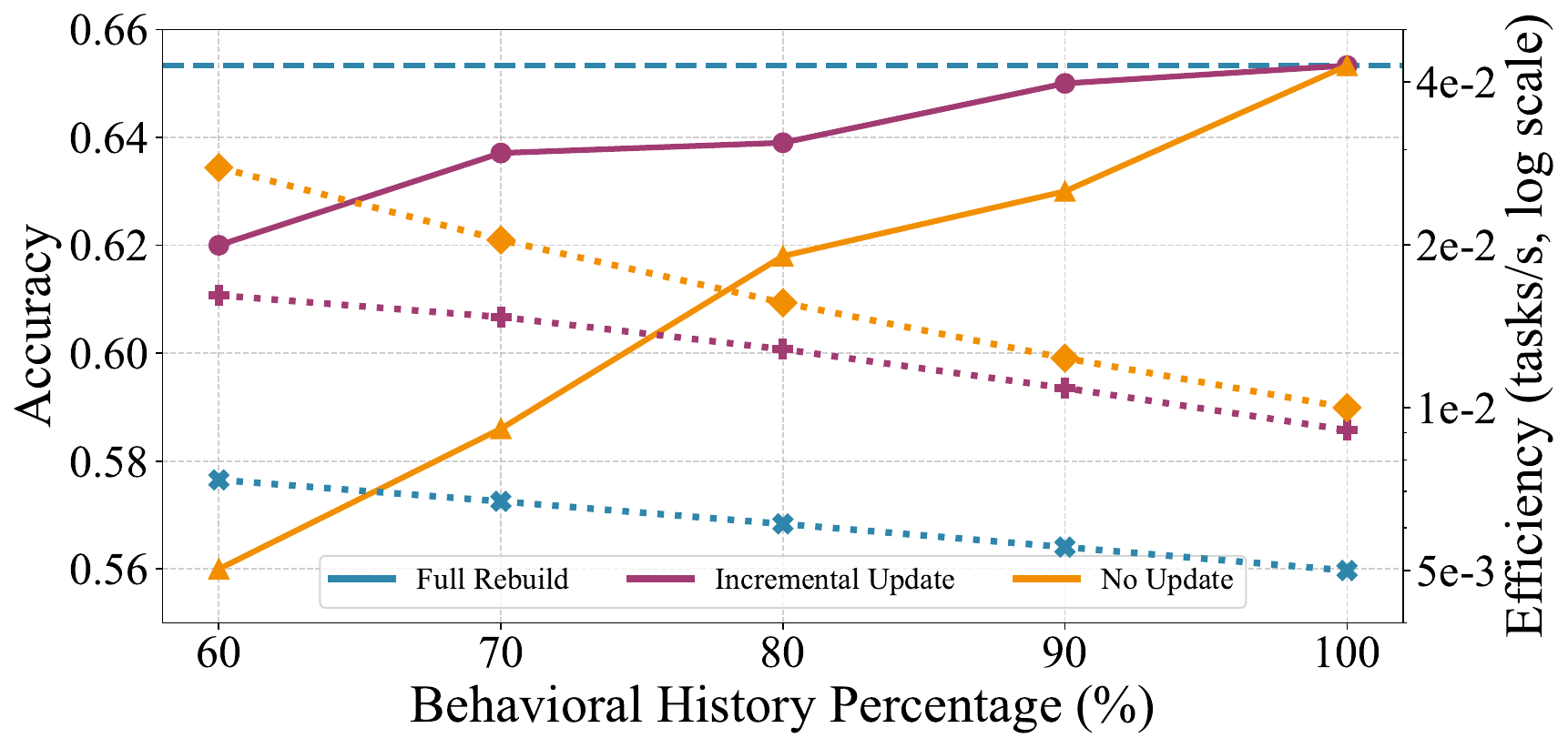}
    \caption{Comparison of memory update strategies in terms of accuracy  and efficiency. }
\label{fig:incremental_update}

\end{figure}
\begin{figure*}[t]
    \centering
    \includegraphics[width=0.98\linewidth]{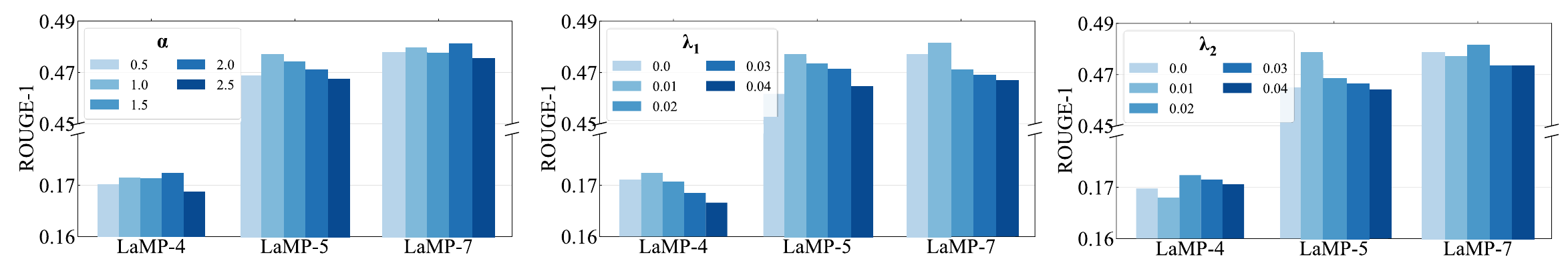}

    \caption{Hyperparameter sensitivity analysis for MemWeaver on the LaMP-4, LaMP-5, and LaMP-7 datasets. The plots show the impact of the semantic guidance weight ($\alpha$), recency bias ($\lambda_1$), and continuity bias ($\lambda_2$) on the ROUGE-1 score. }
    \label{fig:hyper}
\end{figure*}
\subsection{Experimental Analysis}
\subsubsection{Analysis of Edge Traversal Proportions} Figure~\ref{fig:edge_proportion} illustrates the proportion of edges traversed during inference, revealing that the model dynamically adapts to task characteristics. In chronological tasks like news generation, temporal edges constitute the vast majority of paths, while semantic edges gain significant prominence in association-driven domains like citation identification. Crucially, temporal edges account for the majority of traversals across all datasets, validating the fundamental importance of the temporal dimension often neglected by existing methods.
Interestingly, while temporal edges are traversed more frequently, removing semantic edges incurs a larger performance penalty. This disparity suggests that temporal edges form the high-frequency backbone of user behavior, whereas semantic edges act as essential bridges connecting conceptually related but temporally distant activities. Consequently, removing these vital semantic links compromises the graph's structural integrity more severely.

\subsubsection{Analysis of Incremental Update Strategy} To rigorously evaluate the practicality and scalability of our framework in dynamic environments, we designed a simulation experiment modeling the sequential arrival of user interaction data. As illustrated in Figure~\ref{fig:incremental_update}, we benchmark our proposed incremental update strategy against two boundary conditions: a Full Rebuild approach and a No Update baseline. The Full Rebuild method establishes the theoretical accuracy ceiling by utilizing the complete data context for every update; however, its prohibitive computational cost renders it unsustainable for real-time applications. Conversely, the No Update strategy represents the efficiency ceiling but suffers from severe performance degradation, as it fails to capture evolving user interests with a stale profile. Our incremental strategy successfully navigates this trade-off, offering a robust solution that combines the strengths of both extremes. In terms of accuracy, it closely tracks the optimal performance of the Full Rebuild, with any minor initial gaps diminishing rapidly as the user history matures and the graph becomes more complete. Crucially, this high fidelity is achieved with a computational overhead that remains minimal, comparable to the near-instantaneous No Update baseline. These results confirm that our mechanism enables continuous user profile refinement without the latency bottlenecks of graph reconstruction, making it highly suitable for streaming personalization systems.

\subsection{Case Study}
To qualitatively demonstrate the effectiveness of MemWeaver, we present an illustrative case study in Figure \ref{fig:case_study}. The user's behavioral history reveals a pattern of interest in artisanal and experience-oriented hobbies, such as vintage jazz, specialty coffee brewing methods, and DIY crafts. Notably, the user explicitly states a belief that a great gift is about the experience, not the object. Our framework processes this history to construct two forms of memory: a behavioral memory that captures specific, high-relevance interactions, and a cognitive memory that synthesizes a high-level user profile, identifying them as a connoisseur of artisanal hobbies who values experience over material worth. When presented with a query for a unique birthday gift, a standard LLM without memory provides generic suggestions like a food basket or a book. In contrast, the response enhanced by MemWeaver is deeply personalized. 
By recommending a siphon brewing kit aligned with the user's beliefs and artisanal interests, our dual-memory system moves beyond generic suggestions to deliver a response that resonates with the user's core values and latent preferences.

\subsection{Hyperparameter Sensitivity Analysis}

\subsubsection{Impact of Query-Semantic Guidance}
The hyperparameter $\alpha$ controls the degree to which the memory retrieval process is guided by semantic relevance to the current query. This parameter fundamentally balances between a focused, relevance-driven exploitation of the user's history and a broader exploration. Our findings show that performance is suboptimal at both extremes. When semantic guidance is negligible, the random walk becomes untethered from the immediate task, failing to prioritize the most pertinent memories. Conversely, an overly strong semantic guidance leads to a greedy retrieval strategy. This myopic focus on a few highly similar past interactions causes the model to overlook other, more diverse memories that could provide a more holistic context. The optimal performance is achieved with a moderate level of semantic guidance, demonstrating that the most effective strategy is a synergistic one, where the query serves as a compass to direct exploration without overly constraining the path.

\subsubsection{Impact of Recency and Continuity Biases}
We jointly analyze the impact of the recency bias ($\lambda_1$) and the local continuity bias ($\lambda_2$). The former coefficient introduces a preference for temporally recent interactions, while the latter encourages the sampling of sequentially adjacent behaviors from the user's history. Our analysis reveals that the optimal setting for these biases is highly task-dependent, as both excessively high and negligibly low values can be suboptimal. For instance, in scholarly title generation (LaMP-5), which often require a coherent narrative, a moderate preference for recent or contiguous interactions is effective as it helps maintain a focused and evolving context. In contrast, for tasks like tweet paraphrasing (LaMP-7), where a user's interactions are often topically diverse and non-sequential, a strong bias is overly restrictive. In this case, the effect of temporal coherence on the experimental results is relatively small. Ultimately, these findings demonstrate the key principle that  temporal biases must be adapted to the task-specific structure of user behavior.

\section{Conclusion}
In this study, we propose MemWeaver, a novel framework that addresses the challenge of deep personalization from unstructured user histories by restructuring them into a hierarchical memory. The framework's core innovation is the effective interaction between its two memory components. The behavioral memory grounds the model in specific, contextual actions, and the cognitive memory provides abstract guidance on long-term preferences.  Crucially, the framework also supports efficient incremental updates, ensuring its scalability and responsiveness in dynamic applications. Extensive experiments on the six datasets of the LaMP benchmark demonstrate the efficacy of this approach. We hope our framework offers new inspiration for the future of user behavior modeling.

\section*{Acknowledgments}
This research was supported by grants from the National Natural Science Foundation of China (U25B2072, 62502486), the Key Technologies R \& D Program of Anhui Province (No. 202423k09020039) and the Fundamental Research Funds for the Central Universities.

% \section{Acknowledgments}

% Identification of funding sources and other support, and thanks to
% individuals and groups that assisted in the research and the
% preparation of the work should be included in an acknowledgment
% section, which is placed just before the reference section in your
% document.

% This section has a special environment:
% \begin{verbatim}
%   \begin{acks}
%   ...
%   \end{acks}
% \end{verbatim}
% so that the information contained therein can be more easily collected
% during the article metadata extraction phase, and to ensure
% consistency in the spelling of the section heading.

% Authors should not prepare this section as a numbered or unnumbered {\verb|\section|}; please use the ``{\verb|acks|}'' environment.

% \begin{acks}
% \end{acks}

\bibliographystyle{ACM-Reference-Format}
\bibliography{main}


\appendix

\section{Implementation Details}
\label{sec:implementation_details}
\subsection{Overall Setup}
All the experiments are conducted on a server equipped with 2  Nvidia GeForce RTX 4090 GPUs (24GB memory each). Other configuration includes 2  Intel Xeon Gold 6426Y CPUs, 512GB DDR4 RAM, and   1TB SATA SSD, which is sufficient for all the baselines. Our implementation is based on Python 3.10, PyTorch 2.7.1, and the Hugging Face Transformers library, and vLLM (v0.10.0) for efficient large language model (LLM) inference.

\subsection{LLM Configuration}
For the main comparison results presented in Table \ref{tab:main_result}, we utilized two powerful open-source models: \textbf{Llama-3.1-8B-Instruct} and \textbf{Qwen3-8B}, to demonstrate the broad applicability and effectiveness of our MemWeaver framework. For all subsequent experiments, including the ablation studies and in-depth analyses, we used \textbf{Qwen3-8B} as the backbone large language model (LLM). 
During the generation process, we limited the maximum input context length to \textbf{3000 tokens} and set the maximum number of new tokens for generation to \textbf{64}. We employed a nucleus sampling strategy with a temperature of 0.7 and a top-p of 0.95 to ensure a balance between diversity and coherence in the generated text. To account for the stochasticity introduced by this sampling strategy, all experiments for each model were conducted 5 times with different random seeds. The results presented in the tables are the average of these runs. This repetition provides the basis for the statistical significance tests (t-tests) used in our analysis.

\subsection{Memory Construction}
\begin{itemize}
    \item \textbf{Behavioral Memory}: To construct the behavioral memory, we first encoded each document in the user's history into dense vector representations. We employed powerful \textbf{BGE-M3} as the text embedding model for this task, leveraging its strong capabilities in capturing semantic nuances. Following the encoding, we applied the K-means clustering algorithm to group the user's behaviors into a set of K distinct semantic clusters. The number of clusters K was empirically set to 5. For the context-aware random walk, the maximum number of steps was set to 10. The key hyperparameters controlling the transition score---semantic guidance $\alpha$, recency bias $\lambda_1$, and continuity bias $\lambda_2$---were carefully tuned on the validation set, with default values set to $\alpha=1.5$, $\lambda_1=0.01$, and $\lambda_2=0.02$ respectively.

    \item \textbf{Cognitive Memory}: The hierarchical cognitive memory was constructed using the same backbone LLM used for the main generation task (i.e., Qwen3-8B or Llama-3.1-8B). The process involved first segmenting the user's chronological history into temporally coherent phases, generating a local summary for each phase, and finally synthesizing these local summaries into a single, cohesive global summary that constitutes the user's cognitive memory. The prompts for summarization were designed to guide the LLM to identify overarching themes and articulate the evolution of user preferences over time.
\end{itemize}

% --- 数据集详情 ---
\section{Dataset Details}
\label{sec:appendix_dataset}

In this section, we provide detailed descriptions of the LaMP benchmark datasets used in our experiments. Table \ref{tab:dataStatistics} presents the statistics for each dataset partition. Our proposed MemWeaver framework is training-free; it does not require any task-specific fine-tuning and therefore only utilizes the test set for evaluation. In contrast, certain baseline methods that require optimization, specifically ROPG and CFRAG, were trained on the official training set following their respective methodologies. Following that, we describe each task, including its objective, data source, and the structure of the input/output data. Note that the LaMP-6 dataset (Personalized Email Subject Generation) is not publicly available and was therefore excluded from our experiments.

\begin{table}[h]
    \centering
    \caption{Statistics of the LaMP benchmark datasets. The table shows the number of instances in the train, development (Dev), and test sets for each task.}
    \resizebox{\columnwidth}{!}{
        \begin{tabular}{lcccccc}
            \toprule
            \textbf{Dataset} & \textbf{LaMP-1} & \textbf{LaMP-2} & \textbf{LaMP-3} & \textbf{LaMP-4} & \textbf{LaMP-5} & \textbf{LaMP-7} \\
            \midrule
            \#Train & 6,542 & 5,073 & 20,000 & 12,500 & 14,682 & 13,437 \\
            \#Dev & 1,500 & 1,410 & 2,500 & 1,500 & 1,500 & 1,498 \\
            \#Test & 1,500 & 1,557 & 2,500 & 1,800 & 1,500 & 1,500 \\
            \bottomrule
        \end{tabular}
    }
    \label{tab:dataStatistics}
\end{table}

\begin{table*}[h]
\centering
\caption{Performance on Diverse Benchmarks.}
\label{tab:robustness}
\begin{tabular}{lcccccc} % 修正为 7 列 (Dataset, Metric, + 5个模型)
\toprule
\textbf{Dataset} & \textbf{Metric} & \textbf{Non-P} & \textbf{MemoRAG} & \textbf{Memory-R1} & \textbf{CFRAG} & \textbf{MemWeaver} \\ 
\midrule
OpinionQA & Acc. & 0.352 & 0.397 & 0.423 & 0.452 & \textbf{0.473} \\
AmazonRev & R-1  & 0.221 & 0.263 & 0.246 & 0.286 & \textbf{0.302} \\
ChangeMyV & Avg. & 26.59 & 26.98 & 30.15 & 31.25 & \textbf{35.33} \\ 
\bottomrule
\end{tabular}
\end{table*}

\begin{table*}[h]
\centering
\caption{Comparison with Memory-based Baselines on LaMP.}
\label{tab:memory_baselines}
\begin{tabular}{lcccccc} % 修正为 7 列
\toprule
\textbf{Method} & \textbf{LaMP-1} & \textbf{LaMP-2} & \textbf{LaMP-3} & \textbf{LaMP-4} & \textbf{LaMP-5} & \textbf{LaMP-7} \\ 
 & (Acc) & (Acc) & (MAE) & (R-1) & (R-1) & (R-1) \\ 
\midrule
Memory-R1 & 0.579 & 0.385 & 0.398 & 0.156 & 0.454 & 0.440 \\
MemoRAG   & 0.650 & 0.392 & 0.344 & 0.163 & 0.470 & 0.452 \\
Mem0      & 0.652 & 0.424 & 0.330 & 0.162 & 0.465 & 0.451 \\
\textbf{MemWeaver} & \textbf{0.673} & \textbf{0.463} & \textbf{0.280} & \textbf{0.172} & \textbf{0.475} & \textbf{0.479} \\ 
\bottomrule
\end{tabular}
\end{table*}

\paragraph{LaMP-1: Personalized Citation Identification}
This task is a binary classification problem designed to determine an author's citation preference between two candidate papers, based on their publication history. The data is sourced from the Citation Network Dataset (V14). The user's profile consists of titles and abstracts of papers they have authored. User profiles typically contain around 80 entries, with a majority ranging from 50 to 70.
 The input consists of a query that presents two candidate papers and asks the model to choose the one more likely to be cited by the author. This is accompanied by the author's profile, which contains their publication history. The model is expected to output a single identifier, `[1]` or `[2]`, corresponding to the selected paper.

\paragraph{LaMP-2: Personalized Movie Tagging}
This task requires predicting a tag for a given movie from a list of candidates, based on a user's historical movie tagging behavior. The data originates from the MovieLens and MovieDB datasets. The input includes the movie description and a list of possible tags. The user's profile contains movies they have previously tagged. The model's output should be the single most appropriate tag name.

\paragraph{LaMP-3: Personalized Product Rating}
This is a multi-class classification task where the model predicts a 1-5 star rating for a product review, conditioned on the user's past reviews and ratings. The data is from the Amazon review dataset. The input is the text of a new review. The user's profile contains their historical reviews and corresponding scores. The output is a single digit from 1 to 5.

\paragraph{LaMP-4: Personalized News Headline Generation}
In this text generation task, the model must generate a headline for a news article in a style consistent with the author's previous work. The dataset is derived from HuffPost articles. The input is the body of a news article. The user's profile consists of their past article-headline pairs. The output is a generated string representing the new headline.

\paragraph{LaMP-5: Personalized Scholarly Title Generation}
Similar to news headline generation, this task involves generating an academic title for a paper abstract based on the author's prior publications. The data is from the Citation Network Dataset (V14). The input is the abstract of a new paper, and the user's profile contains their past paper titles and abstracts. The output is the generated title.

\begin{figure}[t]
     \centering
     \subfigure[LaMP-1]{
        \label{fig:n_doc_lamp_1}
        \includegraphics[width=0.475\columnwidth]{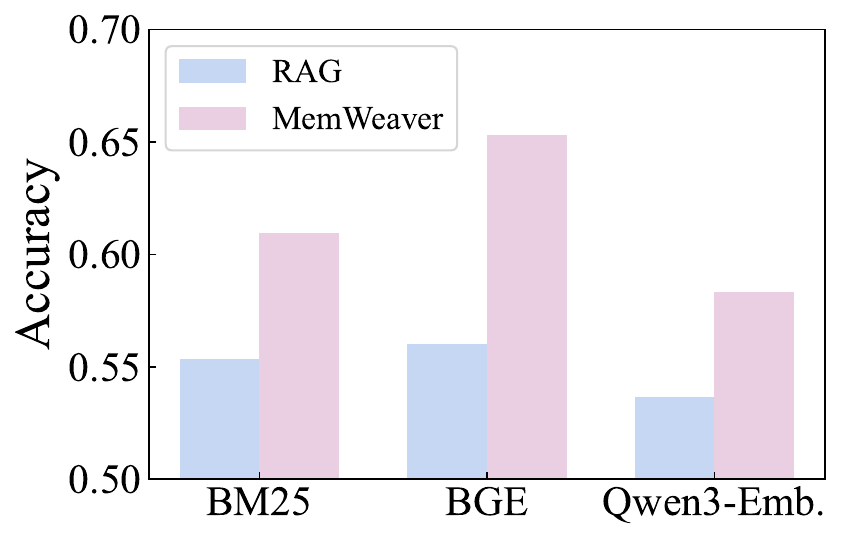}
     }
    \subfigure[LaMP-5]{
        \label{fig:n_doc_lamp_5}
        \includegraphics[width=0.475\columnwidth]{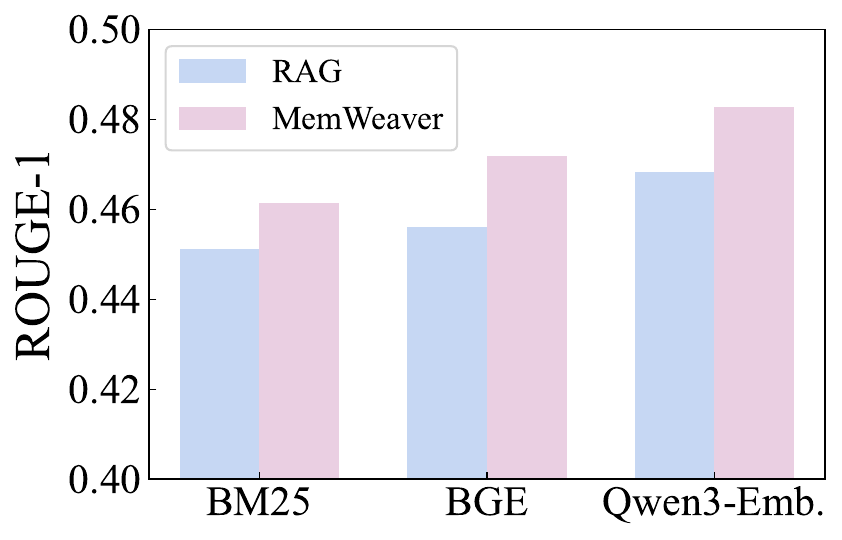}
     }
    \caption{Performance comparison between the baseline RAG and our proposed MemWeaver framework, evaluated with different underlying embedding models.}
    \label{fig:embedding_ablation}
\end{figure}

\begin{figure}[t]
     \centering
     \subfigure[LaMP-1]{
        \label{fig:n_doc_lamp_1}
        \includegraphics[width=0.475\columnwidth]{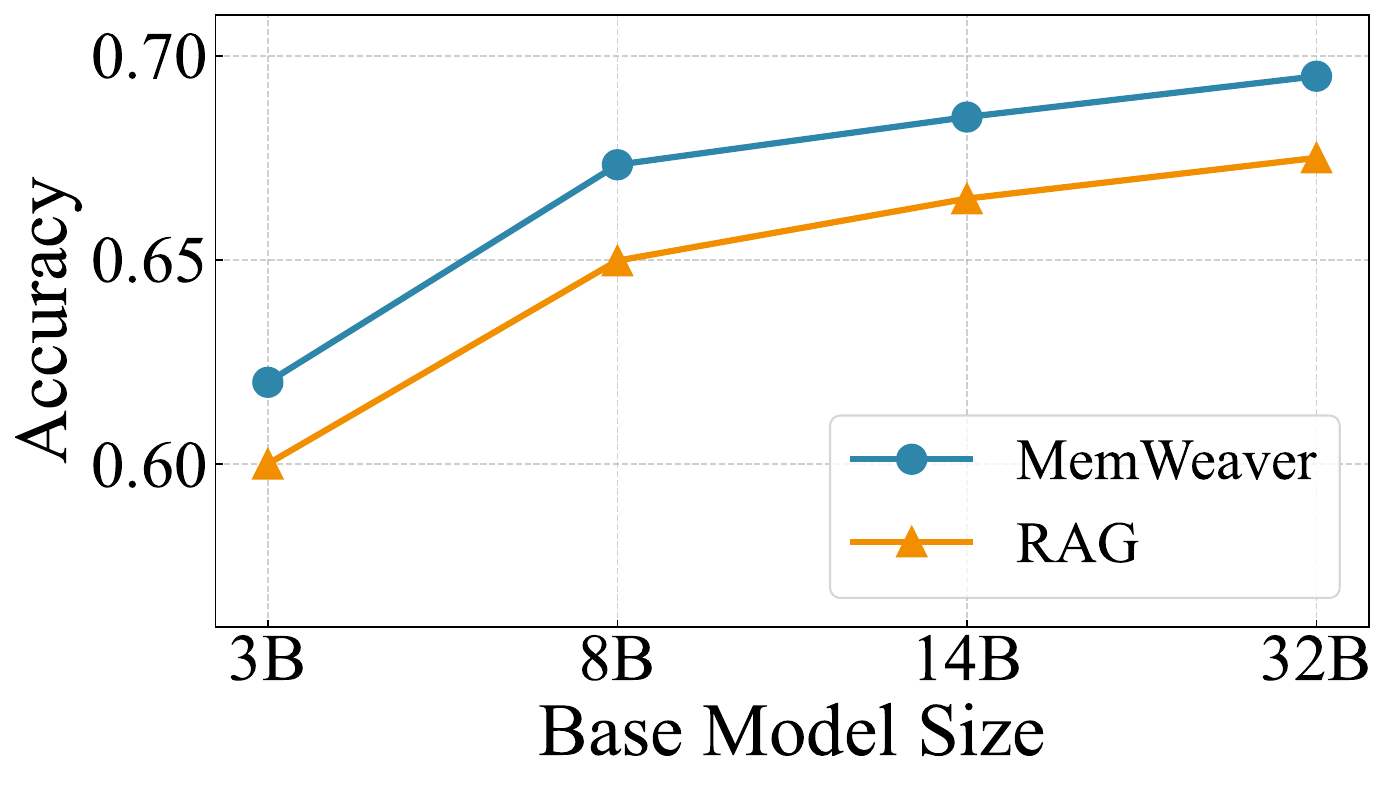}
     }
    \subfigure[LaMP-5]{
        \label{fig:n_doc_lamp_5}
        \includegraphics[width=0.475\columnwidth]{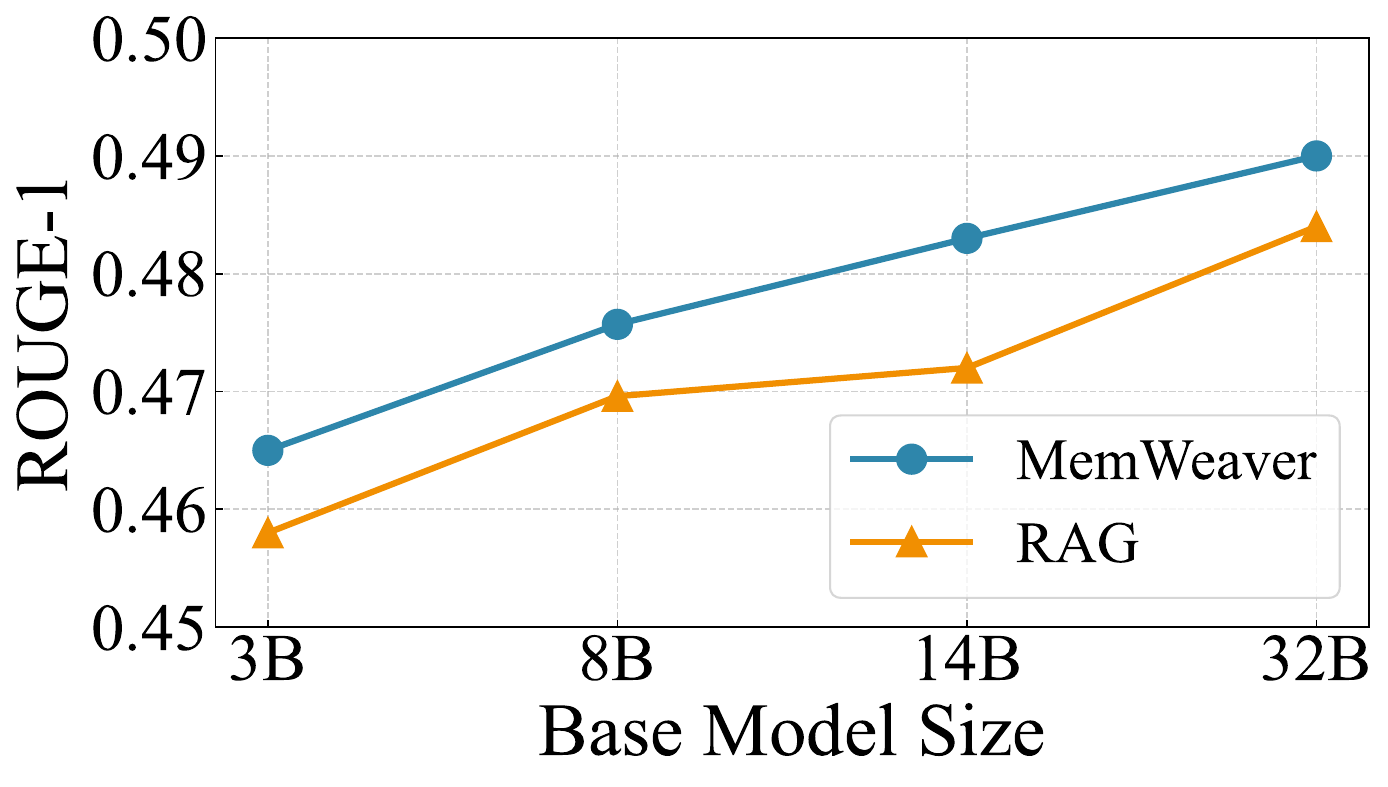}
     }
    \caption{Performance of MemWeaver versus the RAG baseline with increasing base model size. Our method's consistent performance advantage demonstrates its robust improvements with model scaling.}
    \label{fig:scaling_law}
\end{figure}

\begin{table*}[t]
\centering
\caption{Prompt structure for LaMP-1: Personalized Citation Identification.}
\begin{tabular}{lp{0.7\textwidth}}
\toprule
\textbf{Component} & \textbf{Content / Description} \\
\midrule
Cognitive Memory \newline \textit{(Optional)} & A high-level summary of the user's preferences, e.g., "The user is an expert in computer vision and machine learning." \\
\addlinespace
Behavioral Memory \textit{(Optional)} & A list of in-context examples from the user's history, formatted as: \newline \texttt{'title': 'Previous Paper Title 1'} \newline \texttt{'title': 'Previous Paper Title 2'} \\
\addlinespace
Task Instruction & "Based on the historical profiles provided, please choose one of the following two references that is more relevant to the user's input title. Please just answer with '[1]' or '[2]' without explanation." \\
\addlinespace
Query & The candidate references and the current paper's title, formatted as: \newline \texttt{[1] [REFERENCE\_1]} \newline \texttt{[2] [REFERENCE\_2]} \newline \texttt{'title': [QUERY\_TITLE]} \\
\bottomrule
\end{tabular}
\label{tab:prompt}
\end{table*}

\section{Additional Experimental Results}
\label{sec:appendix_exp}
In this section, we provide extended experimental results to further validate the robustness and superiority of the MemWeaver framework. This includes evaluations on diverse benchmarks, comparisons with state-of-the-art memory-augmented LLMs, and sensitivity analyses regarding embedding models and backbone scaling.

\subsection{Robustness Across Diverse Benchmarks} To verify the generalizability of MemWeaver beyond the LaMP benchmark, we extend our evaluation to three additional datasets characterized by distinct personalization challenges: \textbf{OpinionQA} (response prediction)~\cite{personadb}, \textbf{Amazon Reviews} (user-specific sentiment)~\cite{dpl}, and \textbf{ChangeMyView} (persuasive reasoning)~\cite{prime}. As shown in Table~\ref{tab:robustness}, MemWeaver consistently surpasses competitive baselines (including CFRAG and Memory-R1) across all metrics. These results demonstrate that our hierarchical memory structure effectively captures user-specific nuances across various domains, from subjective opinions to stylistic preferences.

% 请确保在文档开头添加：\usepackage{booktabs}

\subsection{Comparison with Memory-based Baselines} We further compare MemWeaver with recent memory-augmented LLM frameworks, including \textbf{MemoRAG}~\cite{memorag}, \textbf{Memory-R1}~\cite{memoryr1}, and \textbf{Mem0}. While these baselines focus on long-context compression or simple key-value storage, they often neglect the structured correlations within user history. As summarized in Table~\ref{tab:memory_baselines}, MemWeaver outperforms these specialized memory models across all LaMP tasks. The performance gap highlights the necessity of our dual-component design, which jointly models behavioral contexts and cognitive abstractions.

\subsection{Impact of Different Embedding Models} To verify that the performance gains of our framework are not solely dependent on a specific embedding model, we evaluate its robustness across three distinct retrieval backbones: a sparse retriever (\textbf{BM25}), a strong dense retriever (\textbf{BGE}), and a large language model-based retriever (\textbf{Qwen3-Embedding}). As illustrated in Figure~\ref{fig:embedding_ablation}, MemWeaver consistently outperforms the standard RAG baseline regardless of the retriever used. This demonstrates that the architectural advantages of our method—namely, its ability to structurally model user history—provide a robust performance lift independent of the underlying retriever's strength.

\subsection{Impact of Base Model Scaling} Finally, we analyze the impact of the base model's scale by evaluating MemWeaver using \textbf{Qwen3-3B, 8B, 14B, and 32B}. As presented in Figure~\ref{fig:scaling_law}, performance on both LaMP-1 and LaMP-5 improves as model size increases. More importantly, MemWeaver maintains a stable and significant advantage over the RAG baseline across all tested scales. This consistency suggests that our structured memory approach provides a foundational enhancement that scales effectively with the base model's inherent capabilities.

\section{Prompt Templates} \label{sec:appendix_prompts}

This section details the prompt structures in MemWeaver. The design prioritizes consistency and interpretability by explicitly structuring roles, contexts, and instructions to standardize how the LLM extracts and applies user-level knowledge.

\subsection{Prompt for Cognitive Memory}

The \textbf{cognitive memory} is constructed via a two-stage summarization pipeline that abstracts long-term user preferences from fine-grained history to coarse-grained semantics.

In the \textbf{Local Summary} stage, the model acts as an \textit{"expert at analyzing user behavior patterns"} to summarize activity clusters (formatted as \texttt{Cluster X Activities (Y records):}). It extracts key themes and local coherence while filtering noise. Subsequently, in the \textbf{Global Summary} stage, the model functions as an \textit{"expert at creating concise user preference summaries."} It aggregates all local summaries (\texttt{Cluster N: [Text]}) into a unified profile (max 300 words). This hierarchical approach ensures semantic generalization by synthesizing specific behaviors into enduring cognitive traits.

\subsection{Prompt for Personalized Generation}

MemWeaver employs a unified \textbf{personalized generation prompt} that integrates two memory sources: \textit{cognitive memory} (high-level global priors) and \textit{behavioral memory} (local, example-level cues).

As shown in Table~\ref{tab:prompt}, this modular structure places both memory components before the task-specific query. By providing complementary signals—global abstractions and concrete evidence—the prompt steers the model toward user-aligned reasoning. This design enhances controllability and transferability across diverse personalized tasks within the LaMP benchmark.

\end{document}